\documentclass[preprint,12pt]{elsarticle}
\biboptions{sort&compress}

\usepackage{amssymb}
\usepackage{amsmath}

\usepackage{xcolor}

\usepackage{todonotes}

\usepackage{amsmath}

\newcommand{\dnthreshold}{\eta}
\newcommand{\riskthreshold}{\theta}

\usepackage{tabularx}
\usepackage{threeparttable}
\usepackage{multirow}

\usepackage{subfig}
\usepackage{newfloat}
\DeclareFloatingEnvironment[fileext=frm,placement={tb},name=Frame]{myFloat}
\usepackage{placeins}

 \usepackage{algorithm}
\usepackage[noend]{algcompatible}
\algrenewcommand\algorithmicindent{2.0em}
\makeatletter
\newlength{\trianglerightwidth}
\algnewcommand{\LineComment}[1]{\State $\triangleright$ #1}
\algnewcommand{\LineCommentCont}[1]{\State \hskip\ALG@thistlm%
  \parbox[t]{\dimexpr\linewidth-\ALG@thistlm}{\hangindent=\trianglerightwidth \hangafter=1 \strut$\triangleright$ #1\strut}}

\usepackage[hyphens]{url}
\usepackage{hyperref}

\usepackage{upgreek}

\usepackage{lipsum}  

\makeatletter
\newcommand\footnoteref[1]{\protected@xdef\@thefnmark{\ref{#1}}\@footnotemark}
\makeatother

\journal{Applied Energy}

\begin{document}

\begin{frontmatter}



\title{Graph Neural Networks for Transmission Grid Topology Control: Busbar Information Asymmetry and Heterogeneous Representations}

\author[radboud,tennet]{Matthijs de Jong \corref{cor1}} 
\author[tennet]{Jan Viebahn } 
\author[radboud]{Yuliya Shapovalova} 
\cortext[cor1]{Corresponding Author: \href{mailto:matthijs.de.jong@tennet.eu}{Matthijs.de.Jong@tennet.eu};
Utrechtseweg 310, Arnhem, 6812 AR, The Netherlands; +31 6 15 01 43 90}

\affiliation[radboud]{organization={Radboud University},
            addressline={Houtlaan 4}, 
            city={Nijmegen},
            postcode={6525 XZ}, 
            country={The Netherlands}}
\affiliation[tennet]{organization={TenneT TSO},
            addressline={Utrechtseweg 310}, 
            city={Arnhem},
            postcode={6812 AR}, 
            country={The Netherlands}}

\begin{abstract}
Factors such as the proliferation of renewable energy and electrification contribute to grid congestion as a pressing problem.
Topology control is an appealing method for relieving congestion, but traditional approaches for topology discovery have proven too slow for practical application.
Recent research has focused on machine learning (ML) as an efficient alternative.
Graph neural networks (GNNs) are particularly well-suited for topology control applications due to their ability to model the graph structure of power grids.
This study investigates the effect of the graph representation on GNN effectiveness for topology control. 
We identify the busbar information asymmetry problem inherent to the popular homogeneous graph representation. 
We propose a heterogeneous graph representation that resolves this problem.
We apply GNNs with both representations and a fully connected neural network (FCNN) baseline on an imitation learning task. 
The models are evaluated by classification accuracy and grid operation ability.
We find that heterogeneous GNNs perform best on in-distribution network configurations, followed by FCNNs, and lastly, homogeneous GNNs.
We also find that both GNN types generalize better to out-of-distribution network configurations than FCNNs.
\end{abstract}

\begin{graphicalabstract}
\end{graphicalabstract}

\begin{highlights}
\item Busbar information asymmetry limits graph neural networks for grid topology control.
\item Heterogeneous graph models of the grid increase model expressiveness and performance.
\item Machine learning agents obtain good grid operation performance at high efficiency.
\item Graph neural networks generalize better to out-of-distribution grid states.
\end{highlights}


\end{frontmatter}





\section{Introduction}

The energy transition, while critical, introduces significant new challenges to transmission system operators (TSOs) \cite{entseo_energy_transistion}.
The inherent variability and limited dispatchability of renewable energy sources can lead to supply peaks.
TenneT, the Dutch TSO, projects an increase in installed renewable power capacity in the Netherlands from 29.6 GW in 2022 to 103.8 GW by 2033 \cite{tennet_2024}.
Simultaneously, the electrification of sectors such as transportation is driving an increase in total electrical demand.
TenneT forecasts Dutch electrical demand to increase from 113 TWh in 2022 to 171 TWh in 3033, representing a growth of over 50\% \cite{tennet_2024}.
Similar trends are anticipated in neighboring countries \cite{tennet_2024}.
These developments introduce challenges such as grid congestion. 

Grid congestion occurs when grid components approach or exceed their operational capacity limits.
Congestion already poses a problem with tangible consequences: the European Union Agency for the Cooperation of Energy Regulators estimated the cost of European grid congestion in 2023 to be approximately 4 billion EUR \cite{ACER2024_CrossZonalCapacities}.
Furthermore, potential grid congestion scenarios can lead to the denial of connection requests for prospective customers, halting the development of businesses, schools, housing, and other facilities.
These problems will quickly escalate without intervention.

Beyond the development of new infrastructure and the rejection of connection requests, TSOs can employ other measures.
TSOs typically use redispatching and/or curtailment, involving the rebalancing or reduction of power production. 
However, both approaches are costly and limited in their applicability.
Topology control presents an alternative method for congestion management that is cost-effective and flexible.
Network topology can be altered by reconfiguring component connections within substations and by disabling power lines.
The resulting change in power flow can alleviate congestion.
Topology control thus present a cheap and flexible congestion management technique \cite{kelly2020rlelec}.

Nevertheless, the usage of topology control by TSOs remains minimal.
The number of possible network topologies grows exponentially with grid size, and most topologies cannot be operated.
Moreover, topological changes affect power flow in complex, nonlinear ways that are time-consuming to compute.
Finding the correct topology for a particular situation is thus difficult, and usage of topology control has hence remained minimal.
However recent advancements have been made with the introduction of GridOptions by TenneT, a pioneering real-world tool for topological strategy suggestion \cite{jan_viebahn_2024}. 
The tool has demonstrated the value of topology control, showing that even simple topological strategies could reduce line congestion by 10 - 20\%.
The tool currently relies on fast load-flow calculations for screening topologies.
However, the tool's scope remains limited to a small section of the Dutch grid, including only eight substations, and the long inference time limits it to day-ahead application.
To remove these limitations, the developers propose the future use of machine learning to speed up computation. 

The idea of mitigating grid congestion with topology control dates back to the 1980s \cite{4334990}.
The discrete, nonlinear nature of topological actions limits the applicability of direct optimization methods and early research therefore often relied on computationally intensive simulations of topological actions.
To address time constraints, simplified models or solvers such as linearized DC approximations were frequently used \cite{515208, 6939408}.
Advanced methods have used more sophisticated solvers \cite{1525118} or techniques to narrow the set of topological actions considered \cite{7792703}.
Nonetheless, these approaches often rely on inaccurate approximations, and their reliance on simulation renders them infeasible for large-scale real-world usage. 
Other research has focused on mathematical optimization techniques, such as mixed-integer programming (MIP) to approximate optimal busbar configurations.
However, such approaches also prove too slow, with proposed approaches being limited to only line-switching \cite{4492805} or busbar-switching within a single substation \cite{9483070}.

Recently, machine learning (ML) techniques have garnered significant interest within the power systems community. 
ML models are notoriously good at learning complex nonlinear patterns, such as the patterns caused by topological actions.
Furthermore, inference with ML models is quick as it does not require extensive search or optimization.
The Learning to Run a Power Network (L2RPN) competitions \cite{marot2019learning, marot2021learning} and the Grid2Op Python library \cite{grid2op}\footnote{\url{https://grid2op.readthedocs.io/}} played a large part in the popularization of this research area.
The field now has an active research community.

Power grid operation is ultimately a sequential decision-making problem. 
ML research on topology control has largely focused on the reinforcement learning \cite{Subramanian_2021, 10.1609/aaai.v37i12.26724} and imitation learning \cite{lan2019aibased, lehna2023managing} paradigms.
Reinforcercement learning (RL) involves training agents by directly interacting with the environment.
However, RL models for topology control often converge to ``do-nothing'' policies because the default topology is mostly satisfactory and topologies that improve upon it are rare \cite{githubGitHubAsprinChinaL2RPN_NIPS_2020_a_PPO_Solution}.
Imitation learning (IL) involves training a ML model to replicate the behavior of an expert agent.
It can be used either as pre-training for RL to avoid the aforementioned problem or as a stand alone method \cite{deJong2024}.
Evolutionary strategies have been also been proposed for the problem \cite{zhou2021actionsetbasedpolicy}.

Graph neural networks (GNNs) \cite{4700287} have recently recieved much attention, both from the machine learning and power engineering community.
Deep learning models have traditionally been designed for tabular, sequential, or grid-like data. 
Consequently, ML methods for power grid operation have historically not or minimally exploited the network structure of power grids.
This is a major limitation as power grid topologies are dynamic and frequently change, for example due to maintenance.
Graph neural networks encode network structure into their representation and are therefore well-suited for power grid modeling.
GNNs have hence been applied to various tasks related to grid operation, such as solving power flows  \cite{8851855, DONON2020106547}, optimal power flow \cite{9053140}, grid reliability or risk assessment \cite{CAMBIERVANNOOTEN2025125401, ZHANG2025124793}, and state estimation \cite{NGO2024122602}.
Similarly, GNNs have been applied to the problem of topology control with promising results \cite{yoon2020winning, 9830198} .
However, despite the existing research, important topics have not yet been addressed by the literature:

\begin{enumerate}
    \item Most present literature applying graph neural networks to topology control uses the same graph representation \cite{9535409, vandersar2023, 9347305, 9830198, Hassouna2025LearningTA}.
    In this representation, configurable objects are connected by edges if they are connected to the same busbar or power line.
    This is, however, only one of many ways to represent the power grid as a graph.
    On the related problem of optimal power flow, progress has been made by considering richer, heterogeneous graph representations \cite{ghamizi2024powerflowmultinet}.
    However, for topology control, a study comparing or proposing improved graph representations for GNNs does not yet exist. 
    \item Power grids are frequently subject to changes, which cannot impede operation. 
    Reliable and robust operation requires models to generalize or adapt to, potentially unexpected,  altered network states.
    Prior work has focused on generalization to out-of-distribution injection profiles \cite{marot2021learning}.
    However, no work has investigated generalization to out-of-distribution network configurations.
\end{enumerate}

This study addresses these gaps in the literature.
We identify a problem with the popular graph representation and propose a heterogeneous graph representation that resolves it.
We train the GNNs, as well as `traditional' fully connected neural networks (FCNNs), on a multi-label binary classification imitation learning task.
The utilized dataset consists of various topologies and outages \cite{deJong2024}. 
The models are evaluated by their classification accuracy and their efficacy when used as agents to operate a simulated power grid. 
Furthermore, we compare the operation efficacy and inference speed of ML models against expert and hybrid agents.
We also investigate the accuracy and operation efficacy of models to out-of-distribution network configurations.
In particular, the foremost contributions of this study are:

\begin{enumerate}
    \item We perform the first analysis of the effect of power grid graph representation on GNN topology control performance.
    The currently popular graph representation models only the current connections between objects.
    However, it ignores which objects might be connected by actions, thereby obscuring the effects of actions.
    We call this the `busbar information asymmetry', which hinders information flow and limits GNN expressiveness.
    We propose a heterogeneous graph representation that models both current and potential connections and consequently resolves the asymmetry.
    We evaluate both graph representations with GNNs.
    \item We perform the first investigation of the ability of graph neural networks to generalize topology control to out-of-distribution grid configurations.
    We evaluate this both by model accuracy and grid operation effectiveness.
\end{enumerate}

This paper is organized as follows. 
Section \ref{sec:power_grid_setup} describes the power grid setup used in this study.
Section \ref{sec:methods} describes the methods used to develop the ML models.
Section \ref{sec:results} describes the evaluation and analysis of the ML models.
Sections \ref{sec:discussion} provides discussion and recommendations for future work, respectively.
The code for this project is available online. \footnote{\label{note_dataset}\url{https://github.com/MatthijsdeJ/GNN_PN_Imitation_Learning}}
Table \ref{tab:notation} lists the meaning of mathematical symbols used in this paper.

\begin{table*}[t]
\centering
\caption{
Mathematical symbols used in this paper and their meaning.
}
\label{tab:notation}
\begin{tabularx}{\textwidth}{lX} 
\hline
\textbf{Notation} & \textbf{Meaning} \\
\hline
$O$ & Set of grid objects (i.e., generators, loads, and line endpoints).  \\ \hline
$S$ & Set of substations. \\ \hline
$\textbf{x}_u$ & Features of a GNN model at node $u \in O$. \\ \hline
$\textbf{h}_{u,k}$ & Node embeddings of a GNN model at node $u \in O$ in layer $k$. \\ \hline
$\mathcal{N}(u)$ & Neighbors of node $u \in O$. \\ \hline
$W$ & Weight matrix.  \\ \hline
$\sigma$ & Activation function.  \\ \hline
$\textbf{p} = (0,1)^{|O|}$ & Output of a ML model.  \\ \hline
$\textbf{y} = \{0,1\}^{|O|}$ & Target of a ML model. \\ \hline
$\alpha = 0.1$ & Label weight hyperparameter. \\ \hline
$\textbf{w} = \{\alpha,1\}^{|O|}$ & Label weights for the loss. \\ \hline
$ \{\textbf{p}_s\} {} \forall s \in S$ & Partition of the output vector $\textbf{p}$ into subvectors $\textbf{p}_s$. 
Subvector $\textbf{p}_s$ corresponds to substation $s \in S$.  \\ \hline
$\dnthreshold$ = 0.97 & Activity threshold parameter. \\ \hline
$\riskthreshold = 1.0$ & Risk threshold parameter, used by various agents. \\ \hline
\end{tabularx}
\end{table*}

\section{Power Grid Setup}
\label{sec:power_grid_setup}

\subsection{Power Grid}

\begin{figure}[tb]
\centering
\includegraphics[width=\textwidth]{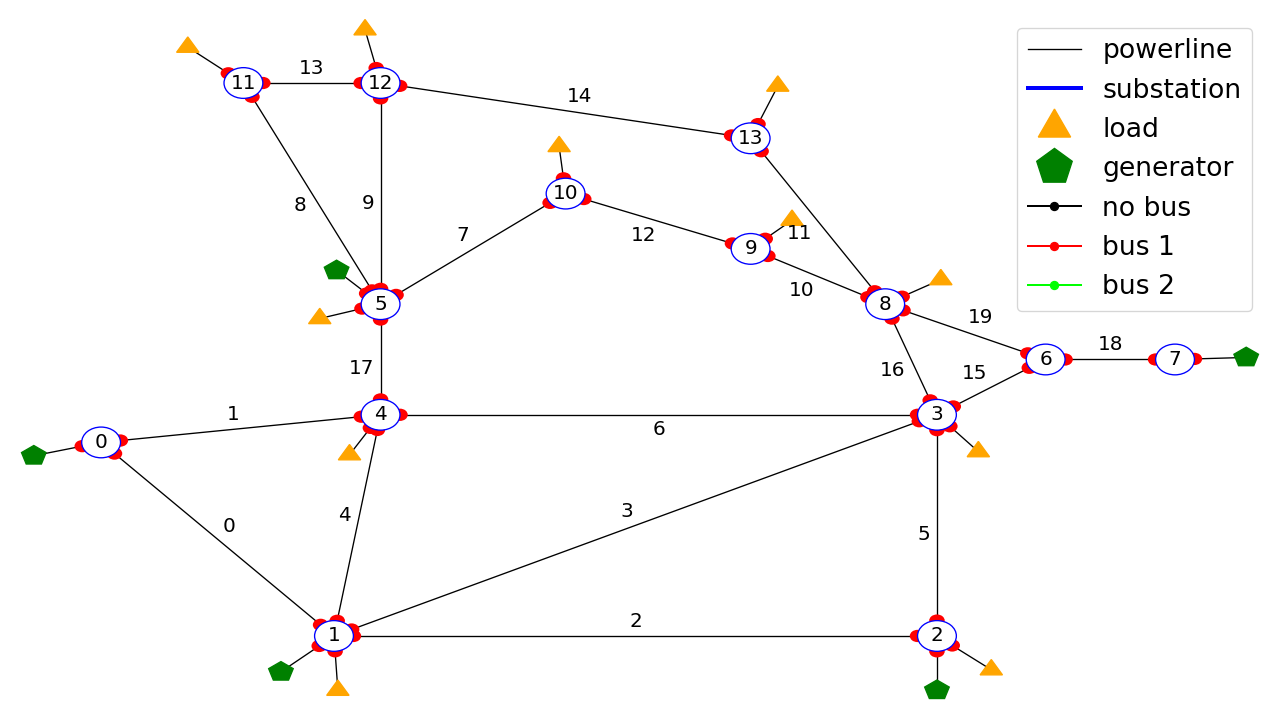}
\caption{The default state of Grid2Op environment \texttt{rte\_case14\_realistic}.
Numbers indicate power line or substation IDs.
}
\label{fig:rte_case14_realistic}
\end{figure}

The IEEE 14-bus system (Fig. \ref{fig:rte_case14_realistic}) was used for the experiment in this study.
Simulations were conducted using \texttt{rte\_case14\_realistic} environment of the Grid2Op library\footnote{Note that this environment has since been deprecated. Future research should use environment \texttt{l2rpn\_case14\_sandbox}.}.
The system comprises 14 substations, five generators (one solar, one nuclear, one wind-based, and two thermal), 11 loads, and 20 power lines.  
Substations 0-4 represent the high-voltage transmission side, while substations 5-13 represent the low-voltage distribution side, interconnected by transformers modeled by lines 15-19. 
To amplify the differences between transmission and distribution characteristics, thermal limits were adjusted according to the values specified by Subramanian et al. \cite{Subramanian_2021}.

Furthermore, we investigate network variations/configurations with single lines disabled, referred to as \textit{N-1 networks}.
These N-1 networks were used to investigate the ability of agents to operate with line outage conditions and to assess the generalization of ML models to out-of-distribution network configurations.

\subsection{Operational Period}
\label{ssec:day_period}
The environment features 1000 distinct scenarios, each spanning 8064 five-minute timesteps, thus modeling a 28-day period.
Each scenario is characterized by unique injection profiles. 
We split the scenarios into individual days, creating 28,000 episodes.
This day-long period better reflects operational periods, and resetting the topology between days emulates beneficial topology reversal \cite{lehna2023managing}.
A simulation terminates early (``game over'') when the power grid is unable to transport sufficient power from generators to loads\footnote{In three scenarios, the Grid2Op power flow solver failed to converge, not as a result of agent misoperation. 
These scenarios were omitted throughout the study.}.

\subsection{Action Space}
\label{ssec:action_space}

Within the simulation environment, each substation is modeled with two busbars. Topology control actions involve switching network \textit{objects} (generators, loads, and line endpoints) between these busbars. A \textit{topology vector} specifies the busbar attachment of each object; each index in the vector corresponds to a specific object, with a value of 1 indicating connection to the first busbar, 2 indicating connection to the second busbar, and -1 indicating disconnection.

Actions are limited to a single substation per timestep, reflecting real-world operational constraints.
Certain substation configurations are considered invalid (e.g., isolated generators or loads) or redundant (configurations mirrored across busbars). 
The action space is generated by applying the filtering approach developed by Subramanian et al. \cite{Subramanian_2021}. 
The procedure is repeated for each network variation to generate the corresponding action spaces (see Subsection \ref{ssec:dataset}).

\subsection{Regimes}
\label{ssec:scope_regime}
To evaluate performance under varying operational stress, three environmental regimes were considered. 
The \textit{full-network regime} does not involve line outages. 
The \textit{planned-outage regime} incorporates static line outages. 
The \textit{unplanned-outage regime} simulates stochastic line outages using the opponent described by Manczak et al. \cite{manczak2023hierarchical}. 
This opponent disables a randomly selected line for a four-hour duration, twice daily, with a minimum one-hour interval between outages.

\section{Methods}
\label{sec:methods}

\subsection{Datasets}
\label{ssec:dataset}

We use the dataset generated by the \textit{greedy} and \textit{N-1} expert agents, described in our previous work \cite{deJong2024}.
Both agents use extensive load flow simulations to find the optimal action.
The greedy agent selects actions by minimizing the maximum line loading of the grid in its current state, whereas the N-1 agent minimizes the maximum line loading under the disablement of any power line.
The N-1 agent demonstrated superior performance compared to the greedy agent in settings both with and without outages.
Both agents use an activity threshold below which ``do-nothing'' actions are automatically selected.
State-action pairs from such do-nothing actions, or from episodes that the expert agents could not successfully complete, were excluded from the dataset.
Do-nothing actions selected above the activity threshold were included.

We constructed two datasets.
First, an \textit{in-distribution} (ID) dataset, containing network configurations on which the models are trained and evaluated. 
This dataset was used in our previous study \cite{deJong2024}.
This dataset incorporates state-action pairs from both the greedy agent and the N-1 agent.
Specifically, it incorporates state-action pairs from the N-1 agent operating on the full network, and from the greedy agent operating on a subset of N-1 networks (lines 0, 2, 4, 5, 6, and 12 disabled).
These specific N-1 networks were selected because the greedy agent can operate them well \cite{deJong2024}.

Second, an \textit{out-of-distribution} (OOD) dataset, which is used to investigate generalization. 
This dataset comprises data from the greedy agent on the N-1 networks with lines 1 and 3 disabled.  

For both datasets, we split the datapoints into 70/10/20 train/validation/test sets based on scenario. 
The partition sizes are listed in Table \ref{tab:N_datapoints}.
In this paper, when we refer to ``ID networks'' or ``OOD networks'', we refer to the network variations in the ID or OOD datasets, respectively.
Likewise, when we refer to ``ID outages'' or ``OOD outages'', we refer to the outages present in respectively ID networks or OOD networks.

\begin{table}[t]
\centering
\begin{threeparttable}[t]
\setlength{\tabcolsep}{30pt}
\caption{The number of datapoints in the datasets and train/validation/test sets.}
\begin{tabular}{llll}
\hline
& Train  & Val & Test \\ \hline
ID & 196,477 & 26,228 & 59,150 \\ \hline
OOD & 41,788\tnote{a} & 5,690\tnote{a} & 12,724 \\ \hline
\end{tabular}
\tnote{a} These datapoints are used to train a model to estimate an upper bound to generalization (see Sec. \ref{ssec:training}).
\label{tab:N_datapoints}
\end{threeparttable}
\end{table}


\subsection{Datapoints}

Each datapoint includes each object's features, the topology vector, and the expert action.
The features per object type are listed in Table \ref{tab:features}.
Object features are normalized.
If a line is disabled, its endpoint features are zero-imputed, and the corresponding topology vector values are -1.
Actions are converted from a set-action format into a switch-action format.
Thus, actions are represented by a vector with a length of the topology vector.
Each index indicates whether the corresponding object is switched between busbars (a value of 1) or not (a value of 0). 
This presents a \textbf{multi-label binary classification task}.

\begin{table*}[b]
\centering
\caption{Features per object type.}
\label{tab:features}
\begin{tabular}{llll} 
\cline{1-1}\cline{3-4}
\textbf{Generator/Load} & & \multicolumn{2}{l}{\textbf{Line endpoint}} \\ 
\cline{1-1}\cline{3-4}
Active Production/Load &  & Active Power Flow & Current flow \\ 
\cline{1-1}\cline{3-4}
Reactive Production/Load &  & Reactive Power Flow & Loading ($\rho$) \\ 
\cline{1-1}\cline{3-4}
Voltage Magnitude &  & Voltage Magnitude & Thermal Limit \\ 
\cline{1-1}\cline{3-4}
\end{tabular}
\end{table*}

\subsection{FCNN}

The fully connected neural network (FCNN) model consists of an input layer, multiple hidden layers, and an output layer. 
The input vector is constructed by flattening the object features into a single vector and concatenating the topology vector. 
The hidden layers use the ReLU activation function.
The output layer uses the sigmoid activation function to constrain the output to the $(0, 1)$ range.
The output vector has the length of the topology vector, so one value $p_u \in (0, 1)$ is predicted per node $u$. 

\subsection{Homogeneous GNN}

\begin{figure}[!tb]
    \centering
    \subfloat[Example Grid]{
        \label{fig:example_grid}
        \includegraphics[width=6cm]{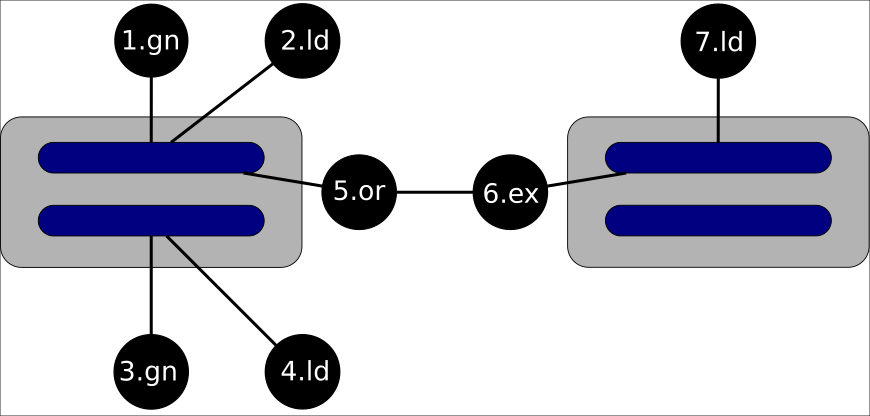}
    }%
    \\ \centering
    \subfloat[Homogeneous Graph Representation]{
        \label{fig:hom_graph_rep}
        \includegraphics[width=6cm]{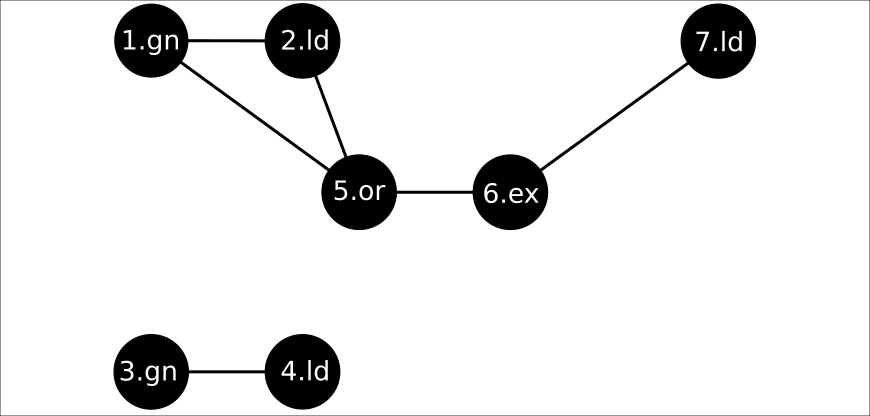}}%
    \subfloat[Heterogeneous Graph Representation]{
        \label{fig:het_graph_rep}
        \includegraphics[width=6cm]{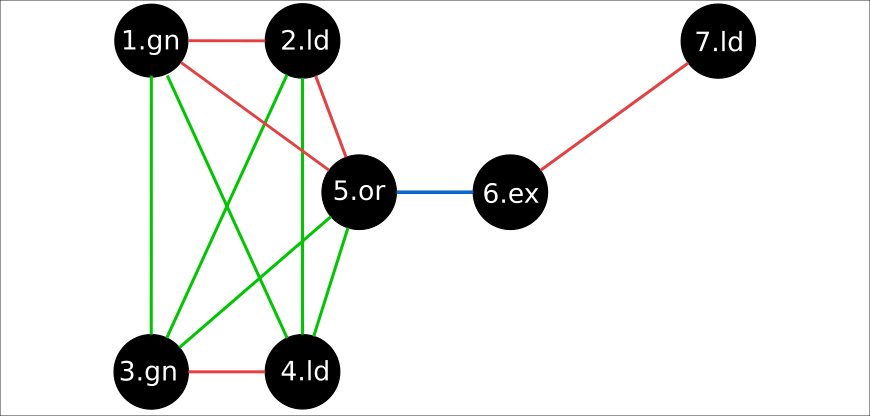}
    }%
    \\
    \subfloat[Homogeneous Message Passing]{
        \label{fig:hom_msg}
        \includegraphics[width=6cm]{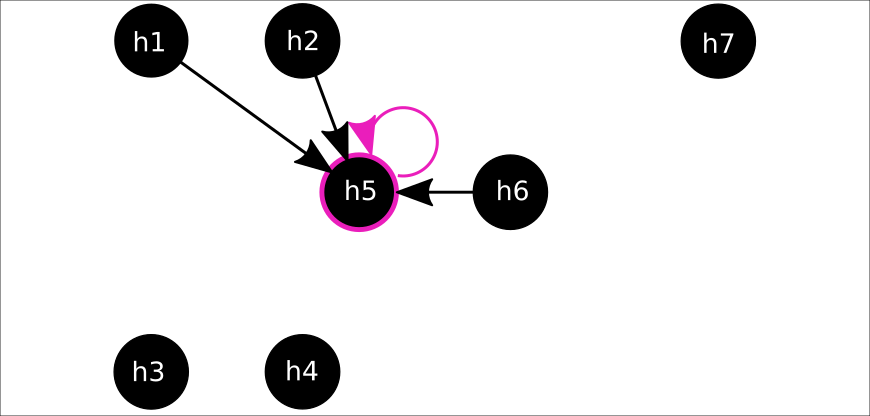}
    }%
    \subfloat[Heterogeneous Message Passing]{
        \label{fig:het_msg}
        \includegraphics[width=6cm]{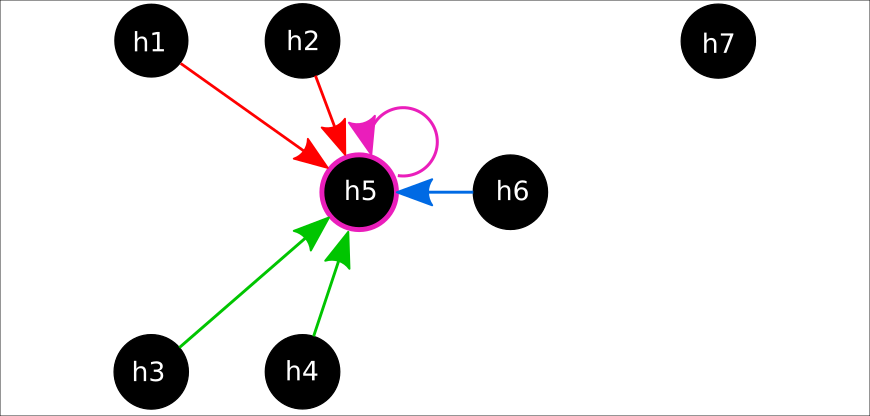}
    }%
    \caption{
        \textbf{a}: An example grid consisting of two substations and seven objects.
        Note that this grid is unrealistic, as real power grids should not be split.
        \textbf{b}: The homogeneous graph representation of that grid.
        \textbf{c}: The heterogeneous graph representation of that grid, where edge colors indicate edge types.
        \textbf{d}: Message passing towards node 5 in the homogeneous model. 
        The purple reflective edge is added to reflect self-weights.
        \textbf{e}: Message passing towards node 5 in the heterogeneous model, where edge colors indicate message types.
    }
    \label{fig:object-busbar&object-object}%
\end{figure}

Applying a graph neural network to the power grid requires a graph representation of the grid.
We represent the grid objects as nodes within the graph.
Edges are set between nodes if the corresponding objects are connected to the same busbar or represent (corresponding) line endpoints.
We call this graph representation \textit{homogeneous} as it does not consider edge types.
We call the associated GNN the \textit{HomGNN}.
Figures \ref{fig:example_grid} and \ref{fig:hom_graph_rep} respectively display an example grid and its homogeneous graph representation.

Different types of grid objects have different features (see Table \ref{tab:features}).
We use multiple two-layer perceptrons to embed the varying object features into a common embedding:
\begin{align*} \textbf{h}_{u,0} = \begin{cases}
\text{MLP}_{gen}(\textbf{x}_u) &\text{node $u$ represents a generator}\\
\text{MLP}_{load}(\textbf{x}_u) &\text{node $u$ represents a load}\\
\text{MLP}_{line}(\textbf{x}_u) &\text{node $u$ represents a line endpoint},
\end{cases}
\end{align*}
where $\textbf{h}_{u,0}$ and $\textbf{x}_u$ are, respectively, the initial node embedding and the features of node $u$.
The embeddings of subsequent layers are computed with the message passing rule \cite{4700287}:
\begin{align}
\textbf{h}_{u,k+1} = \sigma(W_{self,k}\textbf{h}_{u,k} +W_{neighbor,k}\sum_{v\in\mathcal{N}(u)}\textbf{h}_{v,k} + \textbf{b}_k), \nonumber
\end{align}
where $\textbf{h}_{u,k}$ is the embedding for node $u$ in layer $k$, $\sigma$ is the activation function, $W_{self,k}$ and $W_{neighbor,k}$ are the self and neighbor weights in layer $k$, $\mathcal{N}(u)$ is the neighborhood of node $u$, and $\textbf{b}_k$ is the bias term.
Figure \ref{fig:hom_msg} displays an example of homogeneous message passing.
The activation function $\sigma$ is the ReLU function in all but the final layer. 
The final layer uses the sigmoid activation function and outputs one value.
Thus, the model predicts one value $p_u \in (0,1)$ per node $u$.

\subsection{Busbar Information Asymmetry}

The aforementioned graph neural network formulation suffers from a problem.
The output of each node specifies whether to switch the corresponding object across busbars. 
However, messages from objects on the current busbar are passed, while messages from the objects on the other busbar are not.
We call this the \textit{busbar information asymmetry}.
We posit that information about objects on both busbars should be equally available, as both groups of objects are involved in actions.
Moreover, this decreases the model's expressiveness: objects at the same substation but on different busbars and objects at entirely different substations cannot be distinguished.
Lastly, the absence of inter-busbar connections also decreases the graph's connectivity. 
This can result in longer paths that may require deeper GNNs.

\subsection{Heterogeneous GNN}

To address this problem, we propose a \textit{heterogeneous} graph representation, and a corresponding GNN, the \textit{HetGNN}
This heterogeneous representation features three edge types: edges connecting objects on the same busbar, edges connecting objects on the other busbar but within same substation, and edges connecting corresponding line endpoints.
The heterogeneous graph representation is displayed in Figure \ref{fig:het_graph_rep}.
The message passing rule is adapted to consider a weight and neighborhood aggregation per edge type:
\begin{align} \textbf{h}_{u,k+1} = \sigma(&W_{self,k}\textbf{h}_{u,k}
+W_{same,k}\sum_{v\in\mathcal{N}_{same}(u)}\textbf{h}_{v,k}
+ \nonumber \\ 
&W_{other,k}\sum_{v\in\mathcal{N}_{other}(u)}\textbf{h}_{v,k}
+W_{line,k}\textbf{h}_{line,k}
+\textbf{b}_k), \nonumber \label{eqn:diff_neighbors}
\end{align}
where $\mathcal{N}_{same}$ and $\mathcal{N}_{other}$ denote the neighborhoods of objects on the same and other busbar, respectively.
The $W_{line,k}\textbf{h}_{line,k}$ term represents the addition of the embedding of a node connected by a power line.
This term is therefore omitted for nodes that do not represent line endpoints.
Figure \ref{fig:het_msg} displays an example of heterogeneous message passing.

\subsection{Optimization \& Postprocessing}
\label{ssec:setup}

The weights are initialized using a normal distribution, with the standard deviation set as a tuned hyperparameter.
We use the Adam optimizer to minimize a label-weighted binary cross-entropy loss, defined as:
\begin{align*}
L = &\textrm{mean}(-\textbf{w}(\textbf{y}\log(\textbf{p})+  (\textbf{1}-\textbf{y})\log(\textbf{1}-\textbf{p}))), 
\end{align*}
where $\textbf{w}, \textbf{y}$, and $\textbf{p}$ denote the label weight, target, and prediction vectors, respectively.
We introduce label weights to prevent the prediction vectors from collapsing to zeros.
This occurs as the majority of target values are zero.
A lower label weight $0 < \alpha < 1$ is assigned to labels that do not correspond to objects at either the target or predicted substation:
\begin{align}
w_i = \begin{cases}
1 & \text{object $i$ corresponds to the target substation} \\
1 & \text{object $i$ corresponds to the predicted substation} \\
\alpha & \text{otherwise}  
\end{cases} \quad \forall w_i \in \textbf{w}. \nonumber
\end{align}
There is no target substation if the target action is a do-nothing action.
The predicted action is classified as a do-nothing action, without a predicted substation, if all predictions $p_i \in \textbf{p}$ do not exceed $0.5$. 
Otherwise, the predicted substation is the substation where the predictions $\textbf{p}_s$ at that substation $s \in S$ maximize
\begin{align*}
\Sigma_{p_i \in \textbf{p}_s} \textrm{max}(p_i-0.5,0).
\end{align*}

Each value in the prediction vector represents whether to switch the corresponding object.
However, not every prediction vector corresponds to an action in the filtered action space (see Sec. \ref{ssec:action_space}).
We apply a postprocessing step that replaces the model’s prediction $\textbf{p}$ with the nearest action.
This postprocessing step is applied during validation, testing, and inference but not during training.

\subsection{Hyperparameter Tuning \& Training}
\label{ssec:training}

The hyperparameters were tuned with two iterations of hyperparameter sweeps.
This procedure was repeated for the three model types.
The first sweep narrowed the hyperparameter ranges.
It covered wide parameter ranges and trained for a few epochs with strict early stopping.
The second sweep covered narrower ranges and trained with more epochs and less strict early stopping.
All sweeps use random search with Hyperband early termination (distinct from early stopping).
Hyperband early termination terminated unpromising runs early.
Table \ref{tab:hyperparams} describes all final hyperparameter values, the second sweep ranges, and additional clarification where necessary. 

\begin{figure}[tbp]
    \centering
    \includegraphics{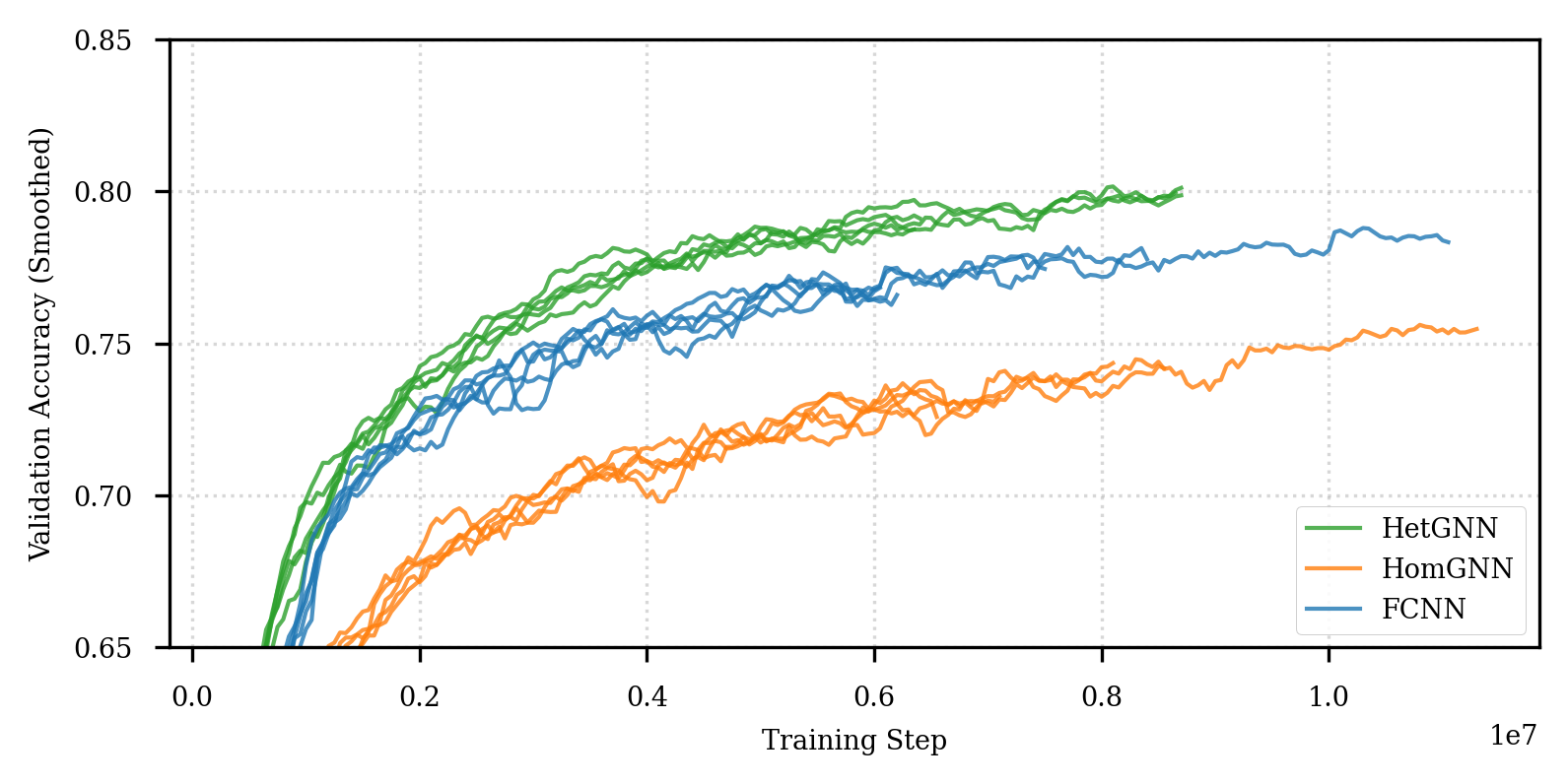}
    \caption{The training curves of the five models per model type.
    }
    \label{fig:training_curves}
\end{figure}

Five models were trained per model type, each with different weight initializations. 
Each training run lasted for 100 epochs unless stopped early.
Runs were terminated early if the highest validation accuracy did not increase in 20 evaluations.
The validation accuracy was calculated every 50,000 iterations.
The training curves are displayed in Figure \ref{fig:training_curves}.
Finally, we trained five heterogeneous GNNs on the OOD data to contrast the models' generalization performance.
These are referred to as \textit{OOD-GNNs}.

\makeatletter
\newcommand{\thickhline}{%
    \noalign {\ifnum 0=`}\fi \hrule height 1pt
    \futurelet \reserved@a \@xhline
}

\begin{table}[tb]
\centering
\setlength{\tabcolsep}{11pt}
\begin{threeparttable}
\caption{Hyperparameters and their final sweep ranges for both the FCNNs and GNNs.}
\label{tab:hyperparams}
\begin{tabular}{lllll}
\hline
Hyperparameter & \begin{tabular}[c]{@{}l@{}} FCNN \\ Value \end{tabular}    & \begin{tabular}[c]{@{}l@{}} FCNN \\ Range \end{tabular} & \begin{tabular}[c]{@{}l@{}}GNN \\ Value\tnote{a} \end{tabular} & \begin{tabular}[c]{@{}l@{}}GNN \\ Range\tnote{a}\end{tabular}  \\ \thickhline
\textbf{\#Hidden Layers}  & 4 & (1, 5) & 8 & (2, 12) \\ \hline
\textbf{\#Hidden Nodes}   & 230        & (32, 256) & 180 & (32, 256) \\ \hline
\textbf{Batch Size} & 64 & (32, 128)  & 64 & (32, 128) \\ \hline
\textbf{Learning Rate}    & 7E-4       & ($e^{-9}, e^{-5})$ & 2E-4 & ($e^{-9}, e^{-5})$ \\ \hline
\textbf{Weight Decay} & 0\tnote{b} & $(e^{-9}, e^{-2})$ & 0\tnote{b} & $(e^{-9}, e^{-2})$ \\ \hline
\textbf{Weight Init. $\sigma$ } & 5          & $(e^{-0.5}, e^{1.5})$ & 5 & $(e^{-0.5}, e^{1.5})$ \\ \hline
\textbf{Label Weight $\alpha$}     & 0.1        & -\tnote{c}          & 0.1       &  -\tnote{c} \\ \hline
\textbf{ReLU neg. slope} \hspace{0cm} & 0.1        & -\tnote{d}          & 0.1       & -\tnote{d} \\ \hline
\end{tabular}
\tnote{a} Hyperparameter tuning was performed independently for both the HomGNN and HetGNN.
However, the results were sufficiently similar.
Hence, the same ranges and final values were selected.
\tnote{b} Weight decay was set to zero after observing that the best runs had values near the lower limit.
Subsequent runs with a weight decay of zero performed better.
\tnote{c} We experimented with label weight $\alpha$ values when introducing label weighting.
Label weights were not included in the hyperparameter sweeps.
\tnote{d} The ReLU negative slope parameter was not tuned.
\end{threeparttable}
\end{table}

\subsection{Imitation Learning Agents}

The ML models are applied to the environment.
They can be applied directly or combined with simulation.
The \textit{naive} agent executes a model's predicted action directly.
We observe that this agent occasionally fails by predicting a singular erroneous action.
The \textit{verify} agent addresses this by verifying predicted actions with simulation.
The predicted action is simulated.
A do-nothing action is selected if the simulated line loadings are increased beyond the thermal limit.
We also consider hybrid agents.
The \textit{verify+greedy} agent normally functions as the verify agent, but switches to the greedy agent if a line breaches the thermal limit.
The \textit{verify+N-1} agent functions similarly but with the N-1 agent.
Each agent also uses the activity threshold used by the expert agents \cite{deJong2024}.

\section{Results}
\label{sec:results}

\subsection{Supervised Learning}

\makeatletter
\renewcommand{\thickhline}{%
    \noalign {\ifnum 0=`}\fi \hrule height 1.2pt
    \futurelet \reserved@a \@xhline
}

\begin{table}[tb]
\centering
\begin{threeparttable}
\caption{The accuracies of the different models on the different combinations of data partitions and ID/OOD network sets. 
Numbers indicate the mean and standard deviation of accuracy for the five models.
Remember that the data partitions are split by scenario.}
\setlength{\tabcolsep}{7pt}
\begin{tabular}{>{\color{gray}\raggedright\arraybackslash}llllll}
\hline
\# & (\%)              & FCNN     & HomGNN & HetGNN & OOD-GNN \\ \thickhline
1 & ID Train        & 79.2±0.9 & 75.2±0.9 & \textbf{83.2}±1.4   & -                                                            \\ \hline
2 & ID Val          & 78.6±0.5 & 75.1±0.8 & \textbf{80.2}±0.6   & -                                                            \\ \hline
3 & ID Test         & 76.6±0.6 & 73.0±0.6 & \textbf{78.5}±0.8   & -                                                            \\ \thickhline
4 & Raw ID Test \tnote{a}        & 73.9±0.6 & 67.9±0.6 & \textbf{74.7}±0.5 & -                                                            \\ \thickhline
5 & Default ID Test \tnote{b} & 94.5±0.9  & 94.2±0.4 & \textbf{95.3}±0.3 & -                                                            \\ \hline
6 & Split ID Test \tnote{c}  & 61.3±0.9 & 54.8±1.2 & \textbf{64.1}±1.2 & -                                                            \\ \thickhline
7 & OOD Train      & 35.0±0.9 & 63.4±1.1 & 67.9±0.7   & \textbf{86.0}±0.6                                                     \\ \hline
8 & OOD Val        & 37.3±0.4 & 65.6±0.6 & 69.3±0.7   & \textbf{83.7}±0.2                                                     \\ \hline
9 & OOD Test       & 34.7±0.7 & 61.1±0.7 & 65.1±0.7   & \textbf{80.7}±0.4                                                     \\ \hline
\end{tabular}
\tnote{a} This lists the accuracy without the postprocessing step described in Sec. \ref{ssec:setup}.
\tnote{b} These datapoints represent the default topology, i.e., without split substations.
\tnote{c} These datapoints represent non-default topologies, i.e., with split substations.
\label{tab:accuracies}
\end{threeparttable}
\end{table}

\begin{figure}[tbp]
     \centering
     \subfloat{\includegraphics{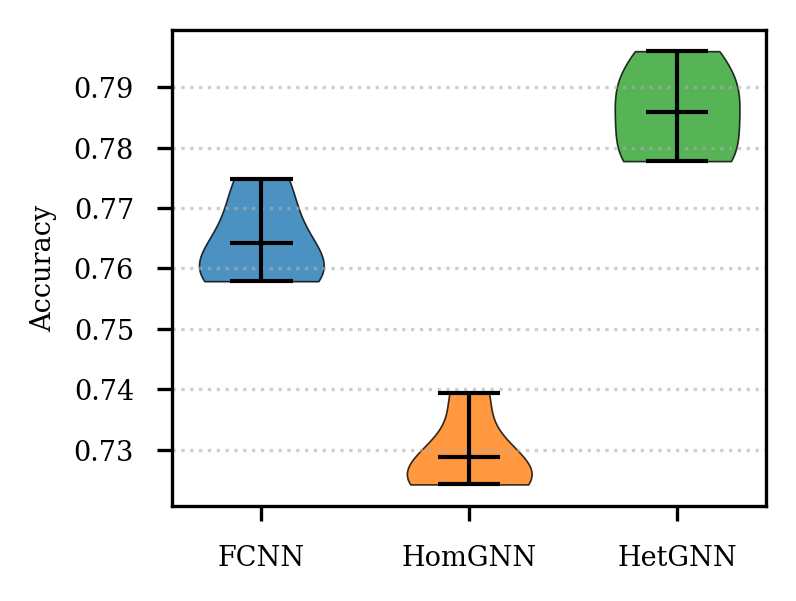}}
     \label{fig:test_accuracies_exposure}%
     \subfloat{\includegraphics{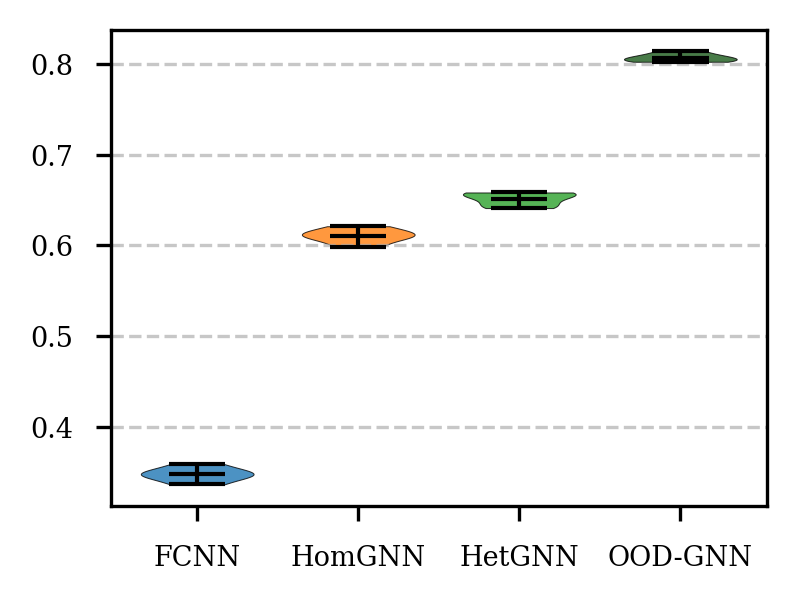}}
     \label{fig:test_accuracies_generalization}
     \caption{
     The test accuracies of the model types on the ID dataset set (left) and OOD dataset (right). 
     The ranges indicate the maximum and minimum accuracy.
     }
     \label{fig:test_accuracies}%
 \end{figure}

Table \ref{tab:accuracies} presents the accuracies of the different models across the data partitions and the ID/OOD network groups.
Figure \ref{fig:test_accuracies} displays the test accuracies of the models on both datasets.
On the ID dataset (rows 1-3), the accuracies remain limited to approximately 80\%. 
On each split, the HetGNN achieves the highest accuracy, followed by the FCNN and, lastly, the HomGNN.
A minor decrease in accuracy is observed between the training, validation, and testing sets for all model types, suggesting a slight degree of overfitting. 
As shown in row 4, accuracy decreases marginally when the post-processing step outlined in Section \ref{ssec:setup} is omitted.

The default topology, characterized by all objects connected to the same busbar, appears frequently within the ID dataset, constituting 46\% of the test set.
This configuration avoids the busbar asymmetry problem, thereby negating the theoretical advantage of the HetGNN over the HomGNN.
As shown in row 5, the accuracies on the standard topology are relatively similar.
Conversely, as shown in row 6, the accuracies vary considerably in topologies with split busbars.

The GNNs achieve far higher accuracies on the OOD dataset than the FCNNs (rows 7-9).
The HetGNNs also achieve higher accuracies than the HomGNNs.
However, the performance of either the HomGNN or the HetGNN on the OOD dataset is still substantially lower than models specifically trained on OOD data (column 'OOD-GNN').

\subsection{Error Analysis}

\begin{figure}[!tb]%
    \centering
    \subfloat{{\includegraphics{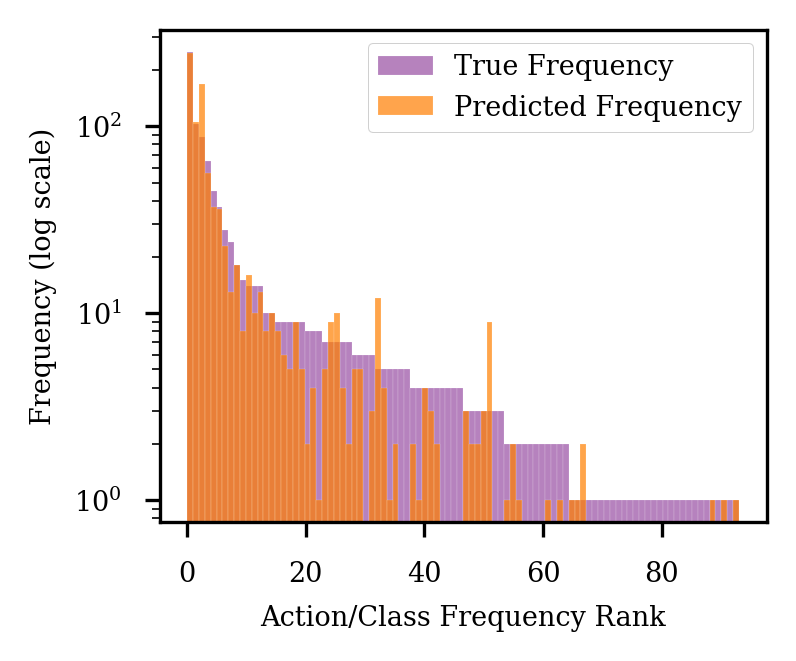}}}
    \subfloat{{\includegraphics{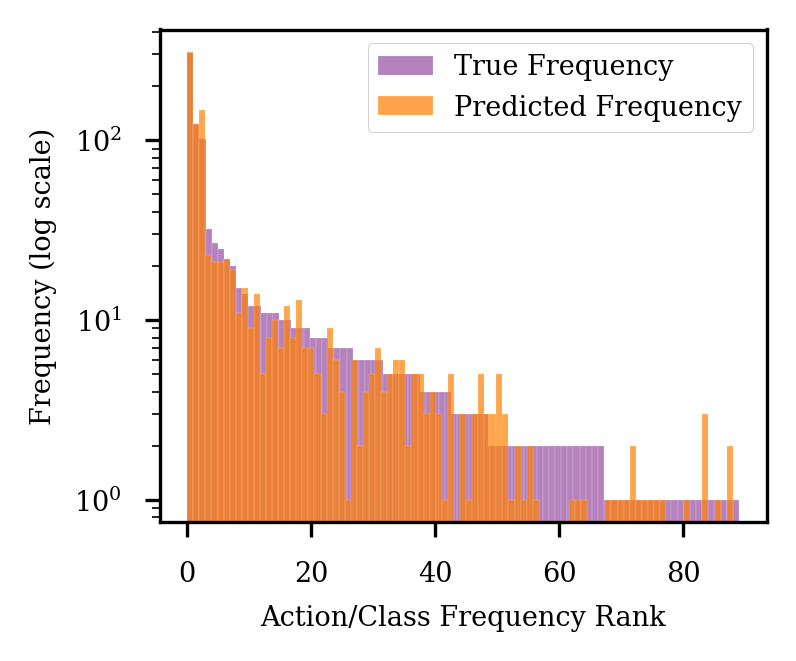} }}%
    \caption{The log-distributions of the classes in the ID validation set.
    Overlaid is the frequency by which the model predicts that class (left: HomGNN, right: HetGNN).
    The non-overlapping blue areas at the tails of the distributions indicate that the models predict rare classes disproportionally infrequently.
    This finding is consistent among the models.}
    \label{fig:label_predictions}
\end{figure}

\begin{figure}[!tb]%
    \centering
    \subfloat{{\includegraphics{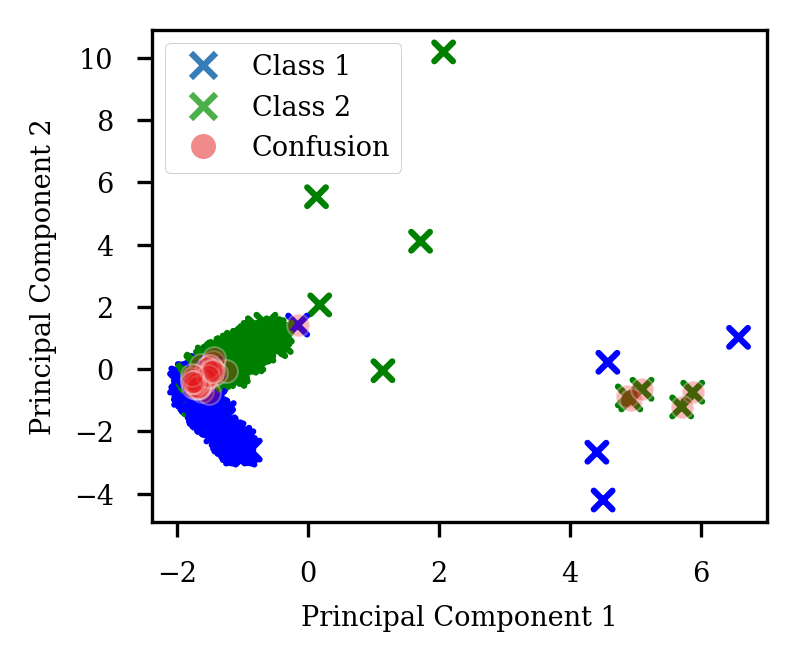}}}
    \subfloat{{\includegraphics{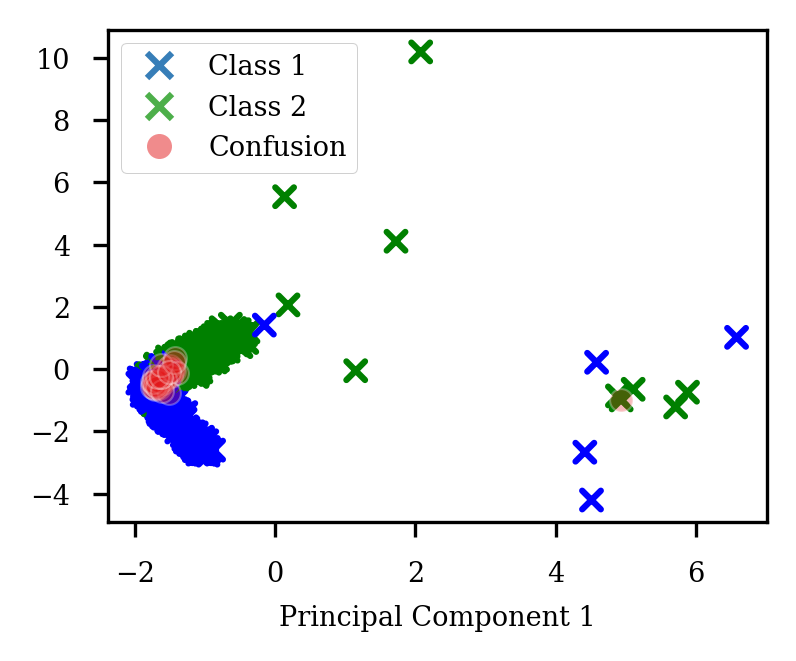} }}%
    \caption{The datapoints of two classes that are most often confused, for the N-1 network with line 0 disabled.
    The datapoints are projected on the first two principal components.
    The datapoints confused by the models (left: HomGNN, right: HetGNN) are overlaid in red.
    As visible, the datapoints in the overlapping region are confused.
    This finding is consistent among the models and confused classes.
    }
    \label{fig:confusions}
\end{figure}

We repeated the error analysis previously performed on the FCNN \cite{deJong2024} for both GNN model types.
As with the FCNN, class imbalance and overlap contribute to the limited accuracy.
As illustrated in Figure \ref{fig:label_predictions}, both GNN types exhibit a tendency to under-predict infrequent classes.
We investigated frequently confused class pairs, i.e. classes commonly mistaken for each other during prediction. 
Figure \ref{fig:confusions} indicates that the errors associated with these frequently confused classes occur within overlapping regions of the feature space.

Analysis of nearest neighbors further supports the hypothesis that class overlap contributes to reduced accuracy.
We applied one of the HetGNNs to a subset of 2,500 validation datapoints.
For data points whose nearest neighbor belongs to the same class, the model achieves an accuracy of 92.97\%. 
In contrast, for data points whose nearest neighbor belongs to a different class, the accuracy drops to 44.44\%. 
Similar trends were observed across models.

\subsection{Graph Smoothness}
\label{ssec:oversmoothing}

Given the well-documented tendency of graph neural networks to exhibit oversmoothing \cite{chen_oversmoothing, 10.5555/3504035.3504468}, which can degrade model performance, we assess smoothness of the graphs produced by both GNNs.
We calculate the mean average distance (MAD) values, which quantify the distance between neighboring node embeddings.
Lower MAD values correspond to higher embedding similarity and thus increased graph smoothness.
Figure \ref{fig:mad} shows the MAD values over the various layers for both GNN types, calculated on the first thousand validation datapoints. 
As shown, the MAD values of the middle GNN layers are consistently higher in the HetGNN than in the HomGNN.
This finding is consistent across the different models per type, suggesting that oversmoothing is a smaller concern for the HetGNN.

\subsection{Graph Diameter}

We mentioned that the heterogeneous graph representation can lead to shorter paths.
We measure this by the diameter, i.e., the shortest path length between the two most distant nodes.
We calculate the diameter by finding the lowest exponent of the adjacency matrix that produces a matrix without zero entries \cite{Duncan2004PowersOT}.
The diameters of the 25 most common topologies are shown in Figure \ref{fig:diameter}.
As visible, the heterogeneous graph representation infrequently leads to a shorter diameter.

\begin{myFloat}
\centering
\begin{minipage}[t]{0.48\textwidth}
\includegraphics{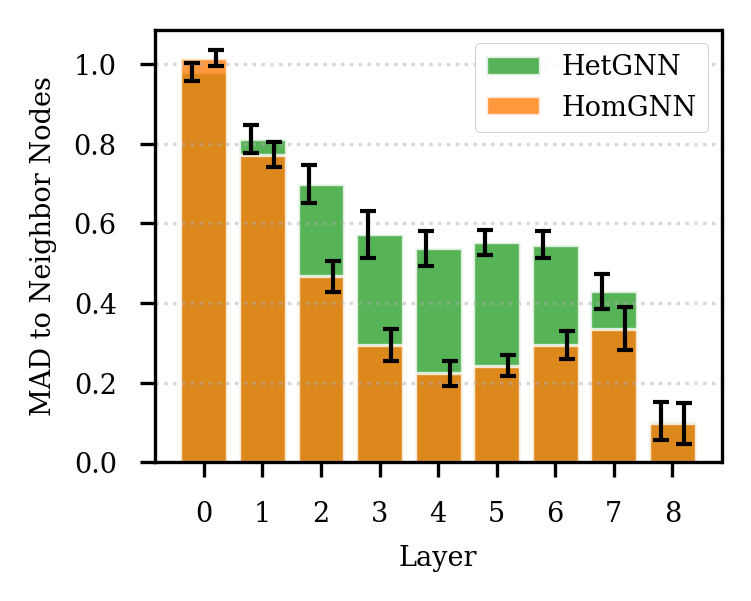}
\centering
  \captionof{figure}{The average neighbor MAD values per layer of a HomGNN and a HetGNN. 
     As visible, the HomGNN has far lower neighbor MAD values in the middle layers.
    Layers 0 and 8 indicate the neighbor MAD values before and after the message-passing layers.
    Error bars indicate the standard deviations over the 1000 datapoints.}
  \label{fig:mad}
\end{minipage}%
\quad
\begin{minipage}[t]{.48\textwidth}
\centering
\includegraphics{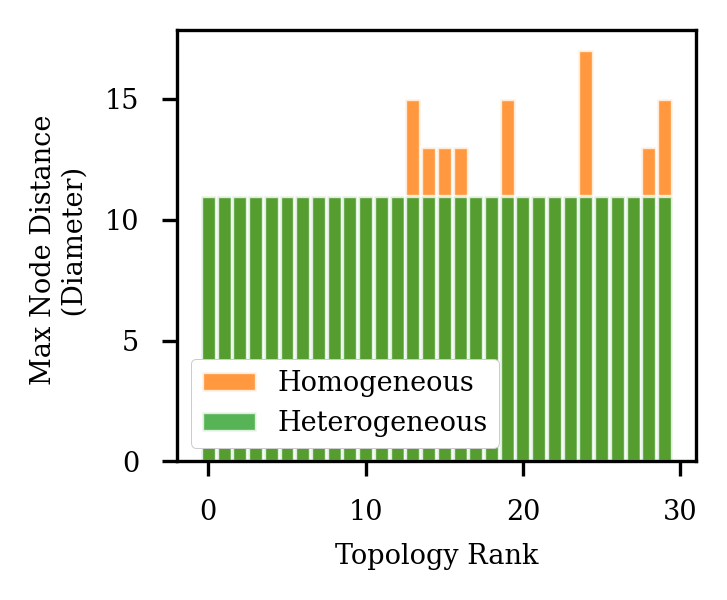}
  \captionof{figure}{The diameter of the 25 most common topologies.
  The diameter of the heterogeneous graph representation is invariant to busbar reconfiguration. }
  \label{fig:diameter}
\end{minipage}
\end{myFloat}

\subsection{Simulation Performance}

\begin{table}[!tb]
\centering
\begin{threeparttable}[tb]
\caption{The percentage of days completed by agents in various regimes.
The agents were evaluated on the test scenarios.
}
\setlength{\tabcolsep}{5.5pt}
\begin{tabular}{>{\color{gray}}llllll}
\hline
\# & \begin{tabular}[c]{@{}l@{}} Agent\\ Type\end{tabular}                      & \begin{tabular}[c]{@{}l@{}}Model\\ Type\end{tabular} & Full Network &  ID Outages\tnote{a} &  OOD Outages\tnote{a} \\ \thickhline
1 & Do-Nothing &  & 59.80 & 46.57±0.17 & 54.20±0.3 \\ \hline
2 & Greedy & & 99.73\tnote{b} & 81.79±1.02 & 86.30±1.12 \\ \hline
3 & N-1 & & 100.00\tnote{b} & 92.39±0.24 & 93.08±0.48 \\ \thickhline
4 & \multirow{4}{*}{Naive\tnote{c}} & FCNN & 96.27±0.90 & 86.33±0.90 & 83.74±1.13  \\ \cline{3-6} 
5 & & HomGNN & 95.39 ± 0.29 & 85.34±0.97 &  86.06±0.60 \\ \cline{3-6} 
6 & & HetGNN & 96.69 ± 0.23 & 87.76±0.57 & 88.02±1.78 \\ \cline{3-6}
7 & & OOD-GNN & - & - &  92.72±0.72 \\ \thickhline
8 & \multirow{4}{*}{Verify\tnote{c}} & FCNN & 98.95±0.19 & 89.62±0.57 & 87.93±0.73 \\ \cline{3-6} 
9 & & HomGNN & 98.82±0.14 & 88.51±0.60 & 89.27±0.80\\ \cline{3-6} 
10 & & HetGNN & 99.35±0.05 & 90.89±0.36 & 91.46±0.86 \\ \cline{3-6}
11 & & OOD-GNN & - & - & 95.01±0.48 \\ \thickhline
12 & \multirow{4}{*}{\begin{tabular}[c]{@{}l@{}} Greedy \\ Hybrid\end{tabular}\tnote{c}} & FCNN & 99.85±0.04 & 93.76±0.16  & 93.23±0.26 \\ \cline{3-6} 
13 & & HomGNN & 99.78±0.05 & 93.44±0.41 & 94.68±0.28 \\ \cline{3-6} 
14 & & HetGNN & 99.88±0.03 & 93.88±0.34 & 94.74±0.67 \\ \cline{3-6}
15 & & OOD-GNN & - & - & 97.82±0.13 \\ \thickhline
16 & \multirow{4}{*}{N-1 Hybrid\tnote{c}} & FCNN & 99.97±0.02 & 95.17±0.18 & 94.49±0.19 \\ \cline{3-6} 
17 & & HomGNN & 100.00±0.00 & 94.72±0.15 & 95.69±0.55\\ \cline{3-6} 
18 & & HetGNN & 100.00±0.00& 95.08±0.35 & 95.41±80 \\ \cline{3-6}
19 & & OOD-GNN & - & - &  98.52±0.11 \\ \hline
\end{tabular}
\tnote{a} These results are averaged over five seeds of outages randomly disabled by an opponent.
\tnote{b} These results were computed over all scenarios as part of data analysis.
\tnote{c} These results are averaged over the five different models.
\tnote{a$\cap$c} In the intersection, results are averaged over five different runs, each with a different model and random outages.
\label{tab:days_completed}
\end{threeparttable}
\end{table}

Table \ref{tab:days_completed} shows the percentage of days completed per agents across the different settings.
The `Full Network' column describes the setting without outages.
The `ID Outages' and `OOD Outages' columns correspond to settings with random outages derived from the in-distribution and out-of-distribution datasets, respectively (see Sec. \ref{ssec:dataset}).

Agent performance follows a consistent order: Do-Nothing $<$ Naive $<$ Verify $<$ Greedy Hybrid $<$ N-1 Hybrid.
The naive agents (rows 4-6) and verify agents (rows 8-10) achieve performance near to the N-1 expert agent (row 3).
The hybrid agents (rows 12-14 and 16-18) can match the performance of the N-1 expert agent.

On the setting with no or ID outages (columns 'Full Network' and 'ID Outages') the HetGNN network consistently performs best, followed by the FCNN and, lastly, the HomGNN (rows 4-6 and 8-10).
Additional simulation reduces or nullifies the differences (rows 12-14 and 16-18).

On the setting with OOD outages (column 'OOD Outages'), the GNNs outperform the FCNN (rows 4-6 and 8-10).
However, additional simulation also reduces this effect (rows 12-14 and 16-18).
Furthermore, the performance of the models trained on the ID dataset remains substantially lower than models directly trained on OOD dataset (rows 7 and 11), even with additional simulation (rows 15 and 19).

\subsection{Computational Efficiency}

We finally consider the inference speed of the different agents.
Inference times are measured using an Apple M1 CPU with minimal background processes across the first fifty validation scenarios. 
Table \ref{tab:durations} lists the mean duration per model and agent type.
Figure \ref{fig:tradeoff} plots these values against the agent's performance on the setting with OOD outages.
The GNNs, particularly the HetGNNs, are considerably slower than the FCNNs.
However, their computation times remain orders of magnitudes smaller than those of the expert agents.
Most importantly, all GNN agents have a favorable combination of speed and performance.

\begin{myFloat}
\centering
\begin{minipage}[t]{0.48\textwidth}
    \vspace*{0pt}
    \includegraphics[width=\linewidth]{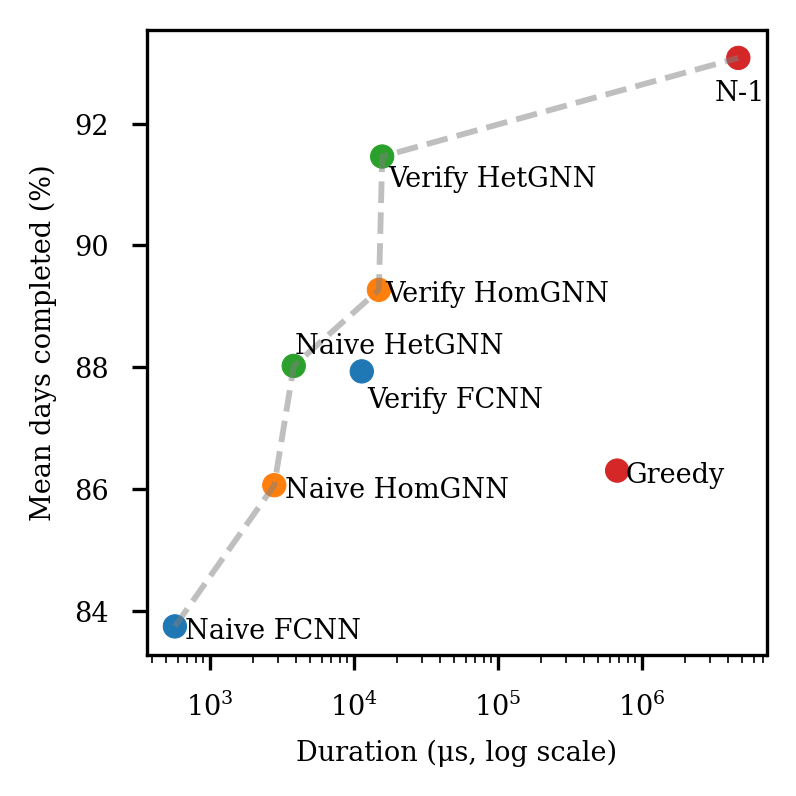}
    \captionof{figure}{The performance and inference duration for the agents on the setting with OOD outages.
    The combinations of speed and performance of the imitation learning agents are favorable.
    All GNN agents are on the Pareto front (dashed line).
    The GNN models are somewhat slower than the FCNN models.}
    \label{fig:tradeoff}
\end{minipage}%
\quad
\begin{minipage}[t]{0.48\textwidth}
\vspace*{0pt}
\setlength{\tabcolsep}{4pt}
\captionof{table}{Inference durations of various agents.
    The hybrid models all use the FCNN.
    The differences between model types for the hybrid agents are relatively insignificant.}
      \begin{tabular}[t]{ll}
    \hline
    \textbf{Agent} & \textbf{Duration ($\upmu$s)} \\ \hline
    Greedy  & 6.77E5 \\ \hline
    N-1  & 4.70E6 \\ \hline
    Naive FCNN  & 5.74E2 \\ \hline
    Naive HomGNN  & 2.81E3 \\ \hline
    Naive HetGNN  & 3.84E3 \\ \hline
    Verify FCNN & 1.14E4 \\ \hline 
    Verify HomGNN & 1.50E4 \\ \hline
    Verify HetGNN & 1.58E4 \\ \hline
    \end{tabular}
  \label{tab:durations}
\end{minipage}
\end{myFloat}

\section{Discussion}
\label{sec:discussion}

\subsection{Interpretation}

The heterogeneous GNN consistently achieves the highest performance, in regards to both classification accuracy and grid operation ability.
The homogeneous GNN exhibits the lowest performance, being surpassed by the FCNN.
The superior performance of the HetGNN compared to the FCNN demonstrates the ability of GNNs to effectively exploit graph structure. 
The inferior performance of the HomGNN can be attributed to the busbar asymmetry problem.
Smoother graphs and longer path lengths might further contribute the inferior performance. 
The more diverse edge representation of the HetGNN likely reduces the oversmoothing effect of GNNs.
Concerning efficiency, the GNNs take considerably longer to evaluate than the FCNNs.
However, this difference is negligible in comparison with the expert agents.

A question that naturally arises is why homogeneous graph neural networks have exhibited good performance in related research \cite{yoon2020winning}. 
We believe that the answer lies in the fact that models used in other research combine GNN layers with other layers (e.g., linear layers), which mitigates the busbar asymmetry problem.
This presents an interesting topic for future work.

GNNs demonstrate a superior ability to generalize to out-of-distribution (OOD) network configurations than FCNNs.
This effect is present in both accuracy and, albeit with reduced magnitude, simulation performance.
Despite their superior generalizability, GNN OOD performance remains substantially below that of models directly trained on OOD network configurations.

The accuracy of the trained models appears to reach a plateau, which may be attributed to issues such as class imbalance and class overlap.
We hypothesize that the observed class overlap originates from the Grid2Op \texttt{simulate} function. 
It is possible that identical network states have diverging forecasts and consequently diverging actions, causing class overlap.
This limitation in classification accuracy does not preclude the models from achieving good operation performance, however.
The naive ML agents achieve operational performance close to the expert agents, whereas the hybrid agents can match them.

Counterintuitively, hybrid agents sometimes achieve performance superior to the expert agents.
A potential source for this effect is the exclusion of samples from the dataset for days that the expert agents could not successfully complete, positively biasing the models to successful behavior.
The model bias towards frequent actions might also result in robust topologies that can accommodate a wider set of injection profiles.

An interesting effect is that model differences are more pronounced with respect to accuracy than operational performance.
Similarly, the limited accuracy does not stop the models from effectively operating the grid.
We propose two contributing factors.
First, a misprediction does not necessarily lead to a game-over; the class overlap suggests that there are often multiple viable actions.
Second, certain challenging days generate a high volume of uncommon data points.
Misclassification of these datapoints affects a considerable subset of the dataset but is limited to only few operational days.

\subsection{Future Work}
\label{sec:future_work}

We propose the integration of the methods developed here, particularly the heterogeneous graph representation, into other approaches as a focus of future research. 
The heterogeneous graph representation could be integrated seamlessly into previously proposed model architectures.
It is particularly still an open question how HetGNN layers interact with non-graphical model layers, warranting investigation.
Incorporating the method into a reinforcement learning framework is also a crucial next step.

It is particularly promising to integrate these approaches into a multi-agent reinforcement learning (MARL) framework, which has recently gained attention for topology control \cite{de_Mol_2025, vandersar2023}.
Within MARL, the combinatorially large action space can be factorized across sub-agents.
The agents developed here could be deployed as localized sub-agents within such a system, offering a natural path toward scaling to larger grids.

Although existing approaches have applied imitation learning only in it's simplest form, there are compelling reasons to apply more advanced methods.
Notably, the simple behavioral cloning learning setup suffers from \textit{distribution shift}, a compounding divergence between the behavior of the ML and expert agent \cite{belkhale2024data}. 
DAgger is a more advanced method that addresses this problem \cite{pmlr-v15-ross11a}.
Additionally, it would also be valuable to further explore expert agents and develop better datasets.

Finally, given the dynamic nature of power grids, generalization of machine learning models remains a topic to be studied more extensively.
There is yet little knowledge about how factors such as the diversity of training data and differences between ID and OOD data affect generalization.
Data augmentation might be a promising technique to obtain models with wide applicability.

\section{Conclusion}

To guarantee energy security, transmission system operators (TSOs) must effectively manage grid congestion.
Topology control guided by machine learning presents a powerful tool for TSOs to do so. 
While graph neural networks (GNNs) are well-suited to model the grid’s structure, we identify that their straightforward application suffers from a critical limitation: the busbar asymmetry problem. 
To address this, we introduce a heterogeneous graph representation that resolves the asymmetry and enables effective learning.
Heterogeneous graph neural networks outperform homogeneous graph neural network, the latter of which fails to outperform the fully-connected neural network (FCNN) baseline.
In turn, heterogeneous GNNs also surpass fully connected neural networks (FCNNs), demonstrating the value of incorporating grid topology explicitly. 
This performance gap is most pronounced in out-of-distribution settings, where FCNN performance deteriorates significantly, while GNNs retain a higher degree of generalization.
Importantly, agents trained with these methods achieve operational performance comparable to expert systems, while offering greatly improved computational efficiency.
These findings demonstrate the potential of GNN-based agents as a solution to support TSOs in real-time grid management.

\section{Statements}

\subsection{CRediT authorship contribution statement}
\textbf{Matthijs de Jong}: Methodology, Software, Formal Analysis, Investigation, Data Curation, Writing - Original Draft, Visualization.
\textbf{Jan Viebahn}: Conceptualization, Methodology, Resources, Writing - Review \& Editing, Supervision, Project Administration.
\textbf{Yuliya Shapovalova}: Methodology, Resources, Writing - Review \& Editing, Supervision.

\subsection{Declaration of competing interest}
Matthijs de Jong reports a relationship with TenneT TSO BV that includes: employment. Jan Viebahn reports a relationship with TenneT TSO BV that includes: employment. If there are other authors, they declare that they have no known competing financial interests or personal relationships that could have appeared to influence the work reported in this paper.

\subsection{Acknowledgements}
Special thanks to Mohamed Hassouna from Fraunhofer Society for providing valuable feedback. 
This work is supported by Graph Neural Networks for Grid Control (GNN4GC) funded by the Federal Ministry for Economic Affairs and Climate Action Germany under the funding code 03EI6117A.

\subsection{Funding}
AI4REALNET has received funding from European Union’s Horizon Europe Research and Innovation programme under the Grant Agreement No 101119527. Views and opinions expressed are however those of the authors only and do not necessarily reflect those of the European Union. Neither the European Union nor the granting authority can be held responsible for them.

\subsection{Code \& Data Availability}
Our code \footnoteref{note_dataset} and data \cite{deJong2024imitation} are publicly available.

\subsection{Declaration of generative AI and AI-assisted technologies in the writing process.}
During the preparation of this work the author(s) used Google Gemini in order to increase the clarity and readability of the text. 
After using this tool/service, the author(s) reviewed and edited the content as needed and take(s) full responsibility for the content of the published article.
The tool was only used for the reviewing and rephrasing of existing text, not the generation of new text.


\begin{thebibliography}{10}
\expandafter\ifx\csname url\endcsname\relax
  \def\url#1{\texttt{#1}}\fi
\expandafter\ifx\csname urlprefix\endcsname\relax\def\urlprefix{URL }\fi
\expandafter\ifx\csname href\endcsname\relax
  \def\href#1#2{#2} \def\path#1{#1}\fi

\bibitem{entseo_energy_transistion}
ENTSO-E, \href{https://eepublicdownloads.entsoe.eu/clean-documents/mc-documents/210414_Financeability.pdf}{European electricity transmission grids and the energy transition} (2021).
\newline\urlprefix\url{https://eepublicdownloads.entsoe.eu/clean-documents/mc-documents/210414_Financeability.pdf}

\bibitem{tennet_2024}
{TenneT TSO B.V.}, \href{https://www.tennet.eu/nl/over-tennet/publicaties/rapport-monitoring-leveringszekerheid}{Rapport monitoring leveringszekerheid}, Tech. rep., TenneT TSO B.V., [Accessed: 25-06-2024] (2024).
\newline\urlprefix\url{https://www.tennet.eu/nl/over-tennet/publicaties/rapport-monitoring-leveringszekerheid}

\bibitem{ACER2024_CrossZonalCapacities}
{European Union Agency for the Cooperation of Energy Regulators (ACER)}, \href{https://www.acer.europa.eu/monitoring/MMR/crosszonal_electricity_trade_capacities_2024}{Capacities for cross-zonal electricity trade and congestion management: 2024 market monitoring report}, Tech. rep., European Union Agency for the Cooperation of Energy Regulators, retrieved March 1, 2025, from \url{https://www.acer.europa.eu/monitoring/MMR/crosszonal_electricity_trade_capacities_2024} (Jul. 2024).
\newline\urlprefix\url{https://www.acer.europa.eu/monitoring/MMR/crosszonal_electricity_trade_capacities_2024}

\bibitem{kelly2020rlelec}
A.~Kelly, A.~O'Sullivan, P.~de~Mars, A.~Marot, Reinforcement learning for electricity network operation (2020).
\newblock \href {http://arxiv.org/abs/2003.07339} {\path{arXiv:2003.07339}}, \href {https://doi.org/10.48550/arXiv.2003.07339} {\path{doi:10.48550/arXiv.2003.07339}}.

\bibitem{jan_viebahn_2024}
J.~Viebahn, S.~Kop, J.~van Dijk, H.~Budaya, M.~Streefland, D.~Barbieri, P.~Champion, M.~Jothy, V.~Renault, S.~Tindemans, \href{https://cse.cigre.org/cse-n035/c2-gridoptions-tool-real-world-day-ahead-congestion-management-using-topological-remedial-actions.html}{{GridOptions} tool: Real-world day-ahead congestion management using topological remedial actions}, {CIGRE Paris} Session 2024 (2024).
\newline\urlprefix\url{https://cse.cigre.org/cse-n035/c2-gridoptions-tool-real-world-day-ahead-congestion-management-using-topological-remedial-actions.html}

\bibitem{4334990}
A.~A. Mazi, B.~F. Wollenberg, M.~H. Hesse, Corrective control of power system flows by line and bus-bar switching, IEEE Transactions on Power Systems 1~(3) (1986) 258--264.
\newblock \href {https://doi.org/10.1109/TPWRS.1986.4334990} {\path{doi:10.1109/TPWRS.1986.4334990}}.

\bibitem{515208}
J.~Wrubel, P.~Rapcienski, K.~Lee, B.~Gisin, G.~Woodzell, Practical experience with corrective switching algorithm for on-line applications, in: Proceedings of Power Industry Computer Applications Conference, 1995, pp. 365--371.
\newblock \href {https://doi.org/10.1109/PICA.1995.515208} {\path{doi:10.1109/PICA.1995.515208}}.

\bibitem{6939408}
M.~Heidarifar, M.~Doostizadeh, H.~Ghasemi, Optimal transmission reconfiguration through line switching and bus splitting, in: 2014 IEEE PES General Meeting | Conference \& Exposition, 2014, pp. 1--5.
\newblock \href {https://doi.org/10.1109/PESGM.2014.6939408} {\path{doi:10.1109/PESGM.2014.6939408}}.

\bibitem{1525118}
W.~Shao, V.~Vittal, Corrective switching algorithm for relieving overloads and voltage violations, IEEE Transactions on Power Systems 20~(4) (2005) 1877--1885.
\newblock \href {https://doi.org/10.1109/TPWRS.2005.857931} {\path{doi:10.1109/TPWRS.2005.857931}}.

\bibitem{7792703}
L.~Wang, H.-D. Chiang, Toward online bus-bar splitting for increasing load margins to static stability limit, IEEE Transactions on Power Systems 32~(5) (2017) 3715--3725.
\newblock \href {https://doi.org/10.1109/TPWRS.2016.2638502} {\path{doi:10.1109/TPWRS.2016.2638502}}.

\bibitem{4492805}
E.~B. Fisher, R.~P. O'Neill, M.~C. Ferris, Optimal transmission switching, IEEE Transactions on Power Systems 23~(3) (2008) 1346--1355.
\newblock \href {https://doi.org/10.1109/TPWRS.2008.922256} {\path{doi:10.1109/TPWRS.2008.922256}}.

\bibitem{9483070}
Y.~Zhou, A.~S. Zamzam, A.~Bernstein, H.~Zhu, Substation-level grid topology optimization using bus splitting, in: 2021 American Control Conference (ACC), 2021, pp. 1--7.
\newblock \href {https://doi.org/10.23919/ACC50511.2021.9483070} {\path{doi:10.23919/ACC50511.2021.9483070}}.

\bibitem{marot2019learning}
A.~Marot, B.~Donnot, C.~Romero, B.~Donon, M.~Lerousseau, L.~Veyrin-Forrer, I.~Guyon, Learning to run a power network challenge for training topology controllers, Electric Power Systems Research 189 (2020) 106635.
\newblock \href {https://doi.org/10.1016/j.epsr.2020.106635} {\path{doi:10.1016/j.epsr.2020.106635}}.

\bibitem{marot2021learning}
A.~Marot, B.~Donnot, G.~Dulac-Arnold, A.~Kelly, A.~O'Sullivan, J.~Viebahn, M.~Awad, I.~Guyon, P.~Panciatici, C.~Romero, Learning to run a power network challenge: a retrospective analysis (03 2021).
\newblock \href {https://doi.org/10.48550/arXiv.2103.03104} {\path{doi:10.48550/arXiv.2103.03104}}.

\bibitem{grid2op}
B.~Donnot, \href{https://GitHub.com/Grid2Op/grid2op}{{Grid2op- A testbed platform to model sequential decision making in power systems. }} (2020).
\newline\urlprefix\url{https://GitHub.com/Grid2Op/grid2op}

\bibitem{Subramanian_2021}
M.~Subramanian, J.~Viebahn, S.~H. Tindemans, B.~Donnot, A.~Marot, Exploring grid topology reconfiguration using a simple deep reinforcement learning approach, in: 2021 IEEE Madrid PowerTech, IEEE, 2021.
\newblock \href {https://doi.org/10.1109/powertech46648.2021.9494879} {\path{doi:10.1109/powertech46648.2021.9494879}}.

\bibitem{10.1609/aaai.v37i12.26724}
A.~Chauhan, M.~Baranwal, A.~Basumatary, {PowRL}: a reinforcement learning framework for robust management of power networks, AAAI'23/IAAI'23/EAAI'23, AAAI Press, 2023.
\newblock \href {https://doi.org/10.1609/aaai.v37i12.26724} {\path{doi:10.1609/aaai.v37i12.26724}}.

\bibitem{lan2019aibased}
T.~Lan, J.~Duan, B.~Zhang, D.~Shi, Z.~Wang, R.~Diao, X.~Zhang, {AI}-based autonomous line flow control via topology adjustment for maximizing time-series {ATCs}, in: 2020 IEEE Power \& Energy Society General Meeting (PESGM), 2020, pp. 1--5.
\newblock \href {https://doi.org/10.1109/PESGM41954.2020.9281518} {\path{doi:10.1109/PESGM41954.2020.9281518}}.

\bibitem{lehna2023managing}
M.~Lehna, J.~Viebahn, A.~Marot, S.~Tomforde, C.~Scholz, Managing power grids through topology actions: A comparative study between advanced rule-based and reinforcement learning agents, Energy and AI 14 (2023) 100276.
\newblock \href {https://doi.org/10.1016/j.egyai.2023.100276} {\path{doi:10.1016/j.egyai.2023.100276}}.

\bibitem{githubGitHubAsprinChinaL2RPN_NIPS_2020_a_PPO_Solution}
{EI Innovation Lab, Huawei Cloud, Huawei Technologies}, {NeurIPS} competition 2020: Learning to run a power network ({L2RPN}) - robustness track, \url{https://github.com/AsprinChina/L2RPN_NIPS_2020_a_PPO_Solution}, [Accessed: 24-06-2024] (2020).

\bibitem{deJong2024}
M.~de~Jong, J.~Viebahn, Y.~Shapovalova, Imitation learning for intra-day power grid operation through topology actions, in: Springer Communications in Computer and Information Science, to appear in ECML-PKDD 2024 post-workshop proceedings. Pre-print available: \url{https://arxiv.org/abs/2407.19865}.

\bibitem{zhou2021actionsetbasedpolicy}
B.~Zhou, H.~Zeng, Y.~Liu, K.~Li, F.~Wang, H.~Tian, Action set based policy optimization for safe power grid management (2021).
\newblock \href {http://arxiv.org/abs/2106.15200} {\path{arXiv:2106.15200}}, \href {https://doi.org/10.48550/arXiv.2106.15200} {\path{doi:10.48550/arXiv.2106.15200}}.

\bibitem{4700287}
F.~Scarselli, M.~Gori, A.~C. Tsoi, M.~Hagenbuchner, G.~Monfardini, The graph neural network model, IEEE Transactions on Neural Networks 20~(1) (2009) 61--80.
\newblock \href {https://doi.org/10.1109/TNN.2008.2005605} {\path{doi:10.1109/TNN.2008.2005605}}.

\bibitem{8851855}
B.~Donon, B.~Donnot, I.~Guyon, A.~Marot, Graph neural solver for power systems, in: 2019 International Joint Conference on Neural Networks (IJCNN), 2019, pp. 1--8.
\newblock \href {https://doi.org/10.1109/IJCNN.2019.8851855} {\path{doi:10.1109/IJCNN.2019.8851855}}.

\bibitem{DONON2020106547}
B.~Donon, R.~Clément, B.~Donnot, A.~Marot, I.~Guyon, M.~Schoenauer, Neural networks for power flow: Graph neural solver, Electric Power Systems Research 189 (2020) 106547.
\newblock \href {https://doi.org/https://doi.org/10.1016/j.epsr.2020.106547} {\path{doi:https://doi.org/10.1016/j.epsr.2020.106547}}.

\bibitem{9053140}
D.~Owerko, F.~Gama, A.~Ribeiro, Optimal power flow using graph neural networks, in: ICASSP 2020 - 2020 IEEE International Conference on Acoustics, Speech and Signal Processing (ICASSP), 2020, pp. 5930--5934.
\newblock \href {https://doi.org/10.1109/ICASSP40776.2020.9053140} {\path{doi:10.1109/ICASSP40776.2020.9053140}}.

\bibitem{CAMBIERVANNOOTEN2025125401}
C.~{Cambier van Nooten}, T.~{van de Poll}, S.~Füllhase, J.~Heres, T.~Heskes, Y.~Shapovalova, \href{https://www.sciencedirect.com/science/article/pii/S030626192500131X}{Graph neural networks for assessing the reliability of the medium-voltage grid}, Applied Energy 384 (2025) 125401.
\newblock \href {https://doi.org/https://doi.org/10.1016/j.apenergy.2025.125401} {\path{doi:https://doi.org/10.1016/j.apenergy.2025.125401}}.
\newline\urlprefix\url{https://www.sciencedirect.com/science/article/pii/S030626192500131X}

\bibitem{ZHANG2025124793}
Y.~Zhang, P.~M. Karve, S.~Mahadevan, \href{https://www.sciencedirect.com/science/article/pii/S0306261924021767}{Graph neural networks for power grid operational risk assessment under evolving unit commitment}, Applied Energy 380 (2025) 124793.
\newblock \href {https://doi.org/https://doi.org/10.1016/j.apenergy.2024.124793} {\path{doi:https://doi.org/10.1016/j.apenergy.2024.124793}}.
\newline\urlprefix\url{https://www.sciencedirect.com/science/article/pii/S0306261924021767}

\bibitem{NGO2024122602}
Q.-H. Ngo, B.~L. Nguyen, T.~V. Vu, J.~Zhang, T.~Ngo, \href{https://www.sciencedirect.com/science/article/pii/S0306261923019669}{Physics-informed graphical neural network for power system state estimation}, Applied Energy 358 (2024) 122602.
\newblock \href {https://doi.org/https://doi.org/10.1016/j.apenergy.2023.122602} {\path{doi:https://doi.org/10.1016/j.apenergy.2023.122602}}.
\newline\urlprefix\url{https://www.sciencedirect.com/science/article/pii/S0306261923019669}

\bibitem{yoon2020winning}
D.~Yoon, S.~Hong, B.-J. Lee, K.-E. Kim, \href{https://openreview.net/forum?id=LmUJqB1Cz8}{Winning the {L2RPN} challenge: Power grid management via semi-markov afterstate actor-critic}, in: International Conference on Learning Representations, 2020.
\newline\urlprefix\url{https://openreview.net/forum?id=LmUJqB1Cz8}

\bibitem{9830198}
S.~Taha, J.~Poland, K.~Knezovic, D.~Shchetinin, Learning to run a power network under varying grid topology, in: 2022 IEEE 7th International Energy Conference (ENERGYCON), 2022, pp. 1--6.
\newblock \href {https://doi.org/10.1109/ENERGYCON53164.2022.9830198} {\path{doi:10.1109/ENERGYCON53164.2022.9830198}}.

\bibitem{9535409}
P.~Xu, J.~Duan, J.~Zhang, Y.~Pei, D.~Shi, Z.~Wang, X.~Dong, Y.~Sun, Active power correction strategies based on deep reinforcement learning—part i: A simulation-driven solution for robustness, CSEE Journal of Power and Energy Systems 8~(4) (2022) 1122--1133.
\newblock \href {https://doi.org/10.17775/CSEEJPES.2020.07090} {\path{doi:10.17775/CSEEJPES.2020.07090}}.

\bibitem{vandersar2023}
E.~van~der Sar, A.~Zocca, S.~Bhulai, \href{https://arxiv.org/abs/2310.02605}{Multi-agent reinforcement learning for power grid topology optimization} (2023).
\newblock \href {http://arxiv.org/abs/2310.02605} {\path{arXiv:2310.02605}}.
\newline\urlprefix\url{https://arxiv.org/abs/2310.02605}

\bibitem{9347305}
P.~Xu, Y.~Pei, X.~Zheng, J.~Zhang, A simulation-constraint graph reinforcement learning method for line flow control, in: 2020 IEEE 4th Conference on Energy Internet and Energy System Integration (EI2), 2020, pp. 319--324.
\newblock \href {https://doi.org/10.1109/EI250167.2020.9347305} {\path{doi:10.1109/EI250167.2020.9347305}}.

\bibitem{Hassouna2025LearningTA}
M.~Hassouna, C.~Holzh{\"u}ter, M.~Lehna, M.~de~Jong, J.~Viebahn, B.~Sick, C.~Scholz, \href{https://api.semanticscholar.org/CorpusID:277112825}{Learning topology actions for power grid control: A graph-based soft-label imitation learning approach}, ArXiv abs/2503.15190 (2025).
\newline\urlprefix\url{https://api.semanticscholar.org/CorpusID:277112825}

\bibitem{ghamizi2024powerflowmultinet}
S.~Ghamizi, J.~Cao, A.~Ma, P.~Rodriguez, \href{https://arxiv.org/abs/2403.00892}{Powerflowmultinet: Multigraph neural networks for unbalanced three-phase distribution systems} (2024).
\newblock \href {http://arxiv.org/abs/2403.00892} {\path{arXiv:2403.00892}}.
\newline\urlprefix\url{https://arxiv.org/abs/2403.00892}

\bibitem{manczak2023hierarchical}
B.~Manczak, J.~Viebahn, H.~van Hoof, Hierarchical reinforcement learning for power network topology control (2023).
\newblock \href {http://arxiv.org/abs/2311.02129} {\path{arXiv:2311.02129}}, \href {https://doi.org/10.48550/arXiv.2311.02129} {\path{doi:10.48550/arXiv.2311.02129}}.

\bibitem{chen_oversmoothing}
D.~Chen, Y.~Lin, W.~Li, P.~Li, J.~Zhou, X.~Sun, Measuring and relieving the over-smoothing problem for graph neural networks from the topological view, Proceedings of the AAAI Conference on Artificial Intelligence 34 (2020) 3438--3445.
\newblock \href {https://doi.org/10.1609/aaai.v34i04.5747} {\path{doi:10.1609/aaai.v34i04.5747}}.

\bibitem{10.5555/3504035.3504468}
Q.~Li, Z.~Han, X.-M. Wu, Deeper insights into graph convolutional networks for semi-supervised learning, in: Proceedings of the Thirty-Second AAAI Conference on Artificial Intelligence and Thirtieth Innovative Applications of Artificial Intelligence Conference and Eighth AAAI Symposium on Educational Advances in Artificial Intelligence, AAAI'18/IAAI'18/EAAI'18, AAAI Press, 2018.
\newblock \href {https://doi.org/10.1609/aaai.v32i1.11604} {\path{doi:10.1609/aaai.v32i1.11604}}.

\bibitem{Duncan2004PowersOT}
A.~J. Duncan, \href{https://api.semanticscholar.org/CorpusID:127195893}{Powers of the adjacency matrix and the walk matrix}, 2004.
\newline\urlprefix\url{https://api.semanticscholar.org/CorpusID:127195893}

\bibitem{de_Mol_2025}
B.~de~Mol, D.~Barbieri, J.~Viebahn, D.~Grossi, \href{http://dx.doi.org/10.1145/3679240.3734602}{Centrally coordinated multi-agent reinforcement learning for power grid topology control}, in: Proceedings of the 16th ACM International Conference on Future and Sustainable Energy Systems, E-Energy ’25, ACM, 2025, p. 460–475.
\newblock \href {https://doi.org/10.1145/3679240.3734602} {\path{doi:10.1145/3679240.3734602}}.
\newline\urlprefix\url{http://dx.doi.org/10.1145/3679240.3734602}

\bibitem{belkhale2024data}
S.~Belkhale, Y.~Cui, D.~Sadigh, Data quality in imitation learning, Advances in Neural Information Processing Systems 36 (2024).

\bibitem{pmlr-v15-ross11a}
S.~Ross, G.~Gordon, D.~Bagnell, \href{https://proceedings.mlr.press/v15/ross11a.html}{A reduction of imitation learning and structured prediction to no-regret online learning}, in: G.~Gordon, D.~Dunson, M.~Dudík (Eds.), Proceedings of the Fourteenth International Conference on Artificial Intelligence and Statistics, Vol.~15 of Proceedings of Machine Learning Research, PMLR, Fort Lauderdale, FL, USA, 2011, pp. 627--635.
\newline\urlprefix\url{https://proceedings.mlr.press/v15/ross11a.html}

\bibitem{deJong2024imitation}
M.~de~Jong, J.~Viebahn, Y.~Shapovalova, \href{https://doi.org/10.17632/w82pscwxgm.1}{Imitation learning topology control dataset} (2024).
\newblock \href {https://doi.org/10.17632/w82pscwxgm.1} {\path{doi:10.17632/w82pscwxgm.1}}.
\newline\urlprefix\url{https://doi.org/10.17632/w82pscwxgm.1}

\end{thebibliography}

\end{document}